%% file: main.tex
\RequirePackage[OT1]{fontenc}
\documentclass[letterpaper, 10 pt, conference]{ieeeconf}
\IEEEoverridecommandlockouts

\usepackage[utf8]{inputenc}
\usepackage{graphics}
\usepackage{epsfig} 
\usepackage{amsmath} 
\usepackage{mathtools}
\usepackage{amsfonts}
\usepackage{amssymb}  
\usepackage{multirow, hhline}
\usepackage{algorithm}
\usepackage{algpseudocode}
\usepackage{lipsum}
\usepackage{subfloat}

\usepackage[dvipsnames]{xcolor}
\usepackage[export]{adjustbox}
\usepackage{soul,color}
\usepackage{mathdots}
\usepackage[font=small]{caption}
\usepackage{subcaption}

\usepackage{textcomp}
\usepackage{gensymb}

\makeatletter
\let\NAT@parse\undefined
\makeatother
\usepackage{hyperref}
\hypersetup{
    colorlinks=true,
    linkcolor=blue,
    filecolor=magenta,      
    urlcolor=blue,
}
\usepackage{cleveref}
\crefname{figure}{Fig.}{Figs.}
\Crefname{figure}{Fig.}{Figs.}
\crefname{section}{Section}{Sections}
\Crefname{section}{Section}{Sections}
\crefname{table}{Table}{Tables}
\Crefname{table}{Table}{Tables}

\newcommand{\eg}[1]{{{\textit{e.g.,}#1}}}
\newcommand{\ie}[1]{{{\textit{i.e.,}#1}}}

\newcommand{\changes}[1]{{{#1}}}

\title{\LARGE \bf
\changes{Topology-Aware Decision Making for Multi-Session \\ Localization and Mapping}
}

\author{Lorenzo~Montano-Oliv\'an*,
        Julio~A.~Placed,
        Luis~Montano,
        and~Mar\'ia~T.~L\'azaro
\thanks{*Corresponding author.}
\thanks{This work was partially supported by DGA\_FSE T73\_23R,
Spanish projects PID2022-\allowbreak139615OBI00/\allowbreak MCIN/\allowbreak AEI/\allowbreak
10.13039/\allowbreak 501100011033/\allowbreak FEDER-UE, and UNDERAIBOT CPP2022-009792/ MICIU/\allowbreak AEI/\allowbreak 10.13039/\allowbreak 501100011033/NextGenerationEU/\allowbreak PRTR, and by EU project MASTERMINE (HORIZON-CL4-2022-RESILIENCE-01, Grant ID: 101091895).
}
\thanks{Lorenzo~Montano-Oliv\'an, Julio A.~Placed, and Mar\'ia T.~L\'azaro are with the Instituto Tecnol\'ogico de Arag\'on (ITA), Mar\'ia de Luna 7, Zaragoza, Spain (e-mail: \{lmontano, jplaced, mtlazaro\}@ita.es).}
\thanks{Luis Montano is with the Instituto de Investigaci\'on en Ingenier\'ia de Arag\'on (I3A), Universidad de Zaragoza, Mar\'ia de Luna 1, Zaragoza, Spain (e-mail: montano@unizar.es).}
}

\usepackage{eso-pic}
\usepackage{mwe}

\begin{document}

\AddToShipoutPictureBG*{%
  \AtPageUpperLeft{%
    \setlength\unitlength{1in}%
    \hspace*{\dimexpr0.5\paperwidth\relax}
    \makebox(0,-0.75)[c]{\parbox{0.9\textwidth}{\centering \small This paper has been accepted for publication in the 2026 IEEE/RSJ Int. Conf. on Intelligent Robots and Systems (IROS). Please cite as: Montano-Oliv\'an, L., Placed, J. A., Montano, L., and L\'azaro, M. T. (2026), Topology-Aware Decision Making for Multi-Session Localization and Mapping. IEEE/RSJ Int. Conf. on Intelligent Robots and Systems (IROS). }}%
}}

\maketitle

\begin{abstract}
\input{abstract.tex}
\end{abstract}

\input{intro.tex}

\input{related_work_iros.tex}

\input{method_iros.tex}

\input{experiments}

\input{conclusions.tex}

\bibliographystyle{IEEEtran}
\bibliography{bibliography}

\end{document}

%% file: abstract.tex
Operating in previously visited environments is becoming increasingly crucial for autonomous systems, with direct applications in autonomous driving, surveying, and warehouse or household robotics.
This repeated exposure to observing the same areas poses significant challenges for mapping and localization
\changes{across sessions, particularly in deciding when a prior model is sufficient for reliable localization and when new mapping is required.}
In this work, we propose a novel multi-session framework that builds on map-based localization, in contrast to the common practice of greedily running full SLAM sessions and trying to find correspondences between the resulting maps.
\changes{The core contribution is a principled, topology-driven mechanism to detect multi-session mapping needs from the pose-graph structure.}
\changes{Specifically, our approach uses spectral connectivity metrics on the joint pose-graph to identify disconnections and weakly constrained regions, and selectively triggers mapping and loop closing only when the pose-graph topology indicates insufficient structural support.}
The resulting map and pose-graph are seamlessly integrated into the existing model, reducing accumulated error and enhancing global consistency \changes{while avoiding redundant remapping}.
We validate our method on overlapping sequences from datasets and demonstrate its effectiveness in a real-world mine-like environment.

%% file: intro.tex
\section{Introduction}

Autonomous operation in previously visited regions is essential to many real-world robotic applications. Examples include indoor service robots deployed in specific buildings, infrastructure inspection over time, autonomous warehouse robotics, and multi-agent exploration.
While Simultaneous Localization and Mapping (SLAM) focuses on building a map and localizing a robot within it during a single traversal of the environment, such applications require enhanced capabilities for repeated operation in the same regions, revisiting some locations and gathering redundant information multiple times.
Multi-session SLAM addresses the challenge of repeated operation in previously visited regions by incrementally building, expanding and refining these maps across sessions. Having accurate, updated maps enables performing downstream tasks more efficiently, and provides the foundation for lifelong SLAM~\cite{kim2022lt}.
In particular, multi-session LiDAR (Light Detection and Ranging) mapping has become increasingly important in several domains where constructing high-precision maps is key; including surveying, search and rescue and autonomous driving~\cite{pang2024lm, hu2024ms}. 

A common approach to multi-session mapping involves executing separate SLAM sessions and subsequently mer\-ging the individual maps; this allows for map growth and consistency~\cite{yin2023automerge}. However, such process involves genera\-ting redundant maps of the same regions and requires an expensive post-processing step to align the full maps.
The online alignment between the prior and current maps of the environment in an incremental fashion has also been introduced to link multiple sessions, typically followed by a Pose-Graph Optimization (PGO) that accommodates the inter-session constraints~\cite{kim2010multiple}.
Nevertheless, these methods still rely on executing time-consuming SLAM algorithms in each session. Moreover, challenges such as drift and insufficient overlap can lead to redundant or inconsistent maps and Pose-Graphs (PGs), and in some cases, unrecoverable states. To reduce computational overhead, the inter-session constraints are often kept sparse, which further limits robustness and scalability.

\begin{figure}
    \centering
    \includegraphics[width=0.9\linewidth]{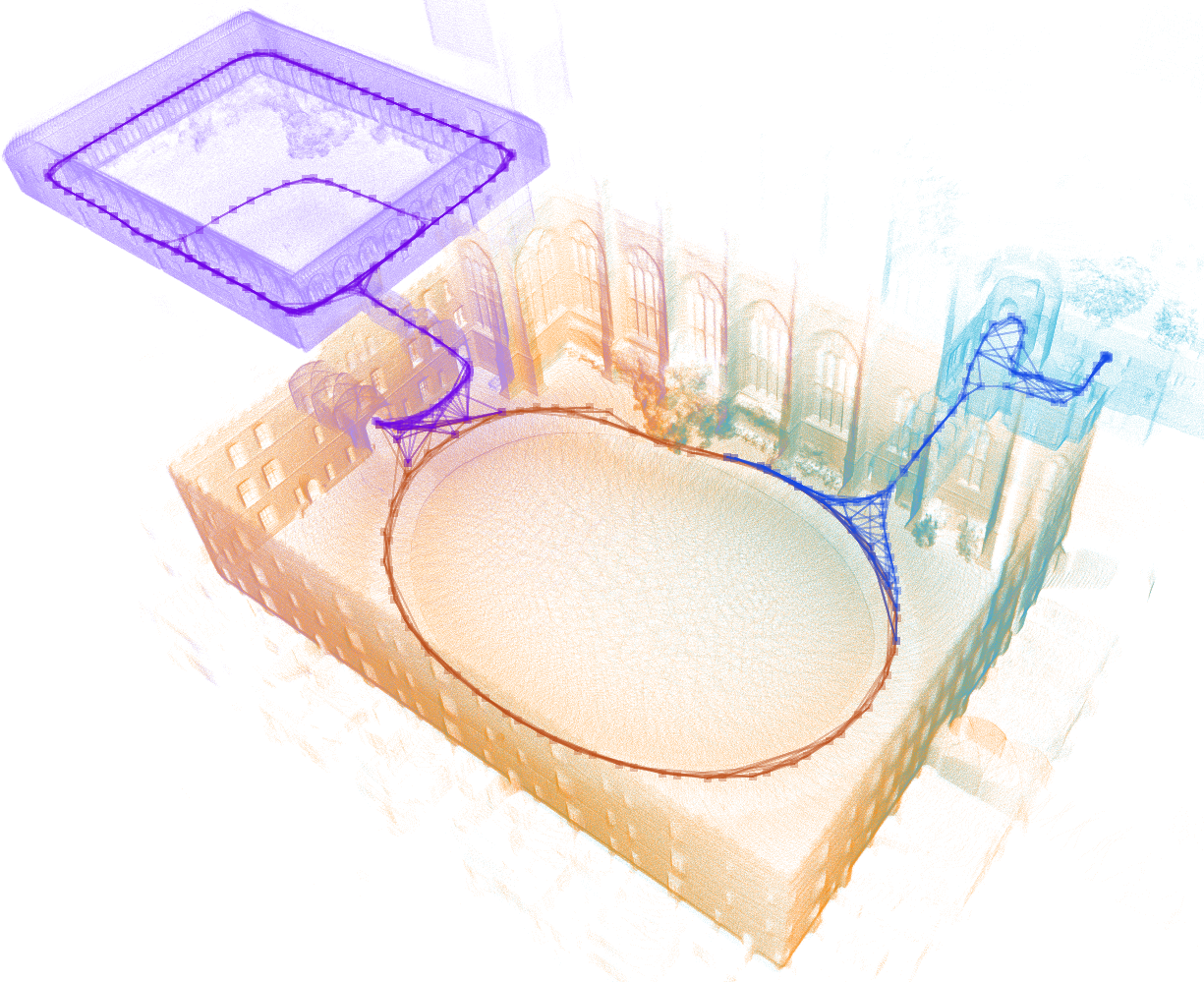}
    \caption{Map and pose-graph built after processing three overlapping sessions of the Newer College Extension dataset~\cite{zhang21} (\emph{Quad M}, \emph{Quad H}, \emph{Cloister}); each depicted in a different color for visualization.}
    \label{fig:nc_3maps}
\end{figure}




Map-based localization provides a \changes{natural} foundation for
\changes{addressing} multi-session mapping. 
Ideally, SLAM would be performed only once to build a high-quality prior map, after which subsequent sessions would rely primarily on localization. In regions not covered by the existing map, mapping and Loop Closing (LC) modules would be selectively activated to expand the environment representation.
Building on this concept, we propose 
\changes{a principled decision-making mechanism that analyzes the topology of the pose-graph to determine whether the current observations are sufficiently supported by the prior model or whether mapping should be triggered. The main contributions are as follows:}
\begin{itemize}
    \item 
    A novel \changes{\textbf{uncertainty-aware decision mechanism} based on spectral graph connectivity metrics, enabling automatic switching between localization and mapping modes.}
    This results in strongly connected graphs across sessions that improve coherency with respect to sparse inter-session loop closures.
    \item
    A \textbf{multi-session system} that integrates such decision-making mechanism \changes{within state-of-the-art map-based localization and LiDAR mapping.} 
    \item
    \changes{Experimental validation} on public datasets \changes{and a real-world underground mine environment, demonstrating that the proposed approach achieves SLAM-level accuracy while reducing unnecessary mapping operations.}
\end{itemize}

%% file: related_work_iros.tex
\section{Related Work}

\subsection{Multi-session Mapping}

Multi-session capabilities have been extensively studied in the visual SLAM literature. 
In~\cite{kim2010multiple}, loop closure factors are employed to jointly optimize multiple graphs (\ie~trajectories) from different sessions.
Daoud \textit{et al.}~\cite{daoud2018slamm} combine multiple maps online by adding these constraints between keyframes with enough similarity.
ORB-SLAM Atlas~\cite{elvira2019orbslam} improves the previous by treating the maps as a single entity with a shared descriptor database. In addition, they remove duplicate points, but map merging requires $\approx13$x frame rate.

In the domain of LiDAR mapping, where the accuracy and consistency of the maps are crucial, multi-session remains an even more challenging problem.
Many works have addressed this problem by creating entirely new maps for each session and combining them either online or offline. 
LT-Mapper~\cite{kim2022lt} combines multiple maps from independent sessions using ScanContext and Iterative Closest Point (ICP) verification. The constraints are included as anchor node-based inter-session loop factors to reduce the optimization complexity, and rely on a robust back-end to mitigate inevitable false loop detections. 
In~\cite{LazaroIROS18}, the optimization complexity across sessions is kept bounded through the use of condensed measurements, which enable the merging of local maps upon detecting inter-session loop closures, while simultaneously removing out-dated nodes during the fusion process.
Automerge~\cite{yin2023automerge} demonstrates large-scale (\ie~hundreds of km) map merging. However, it heavily depends on sufficient overlapping between maps and the availability of Global Navigation Satellite Systems (GNSS) for degraded environments. In addition, the matching complexity scales quadratically with the number of session keyframes, requiring minutes to complete. 
FRAME~\cite{stathoulopoulos2024frame} employs learned descriptors and place recognition to detect overlap between point cloud maps, boosting the map matching process. Despite this is treated as a triggered event, the events are predefined by the user (\eg~when two robots are nearby).
LM-Mapping~\cite{pang2024lm} propose a two-level optimization with local odometry constraints and inter-session factors. Despite this allows to improve local precision and mitigate inter-map inconsistencies, it still requires to rebuild the maps on every session and is relegated to an offline post-processing phase.
In MS-Mapping~\cite{hu2024ms}, each new LiDAR cloud is registered both sequentially and with chunks of the prior map, incrementally building the graph and a merged map. Once again, the multi-session mapping problem is formulated as a PGO, where inter-session constraints are modeled as loop closure factors. This approach is shown to outperform direct map-to-map and \changes{scan}-to-map registration.
LEMON-Mapping~\cite{wang2025lemon} presents a scalable multi-robot system that leverages geometric constraints from overlapping point clouds to strengthen the role of inter-session LC. Reliable loop selection and joint inter-session (hierarchical) PGO are shown to be crucial for mitigating map divergence across sessions.

The proposed system is aligned with~\cite{hu2024ms}. Unlike other related work, we avoid the costly construction, maintenance and matching of complete maps in revisited regions ---we only keep a local \emph{submap}. Instead, we only perform mapping and LC for new regions and ensure strong connections with the existing ones.
Following~\cite{stathoulopoulos2024frame}, we treat this as a triggered event; however, in our case, it is not user-defined but occurs automatically.

\subsection{Map-based Localization}
The problem of multi-session mapping is closely related to the \emph{kidnapped robot problem}~\cite{mcdonald2013real}, where the goal is to estimate the robot's pose given a prior map and no \emph{a priori} information about its location within it.
In such cases, map-based localization emerges as a promising solution, despite its accuracy is limited to prior map precision.
G-Loc~\cite{Montano-OlivanRAL2024} presents a robust PG-based method \changes{for} LiDAR-inertial localization using prior topological and metric information. However, its operation in unmapped regions is limited and it does not incorporate a way to expand the prior map.
LTA-OM~\cite{zou2024lta} first triggers a localization method and then loads the prior map into the odometry module~\cite{xu2022fast}, thus not requiring additional map merging operations nor rebuilding a map when operating in the same area.
In contrast to~\cite{Montano-OlivanRAL2024} and~\cite{zou2024lta}, \changes{which primarily focus on localization using a prior model, this work addresses the complementary challenge of determining when the prior model is no longer sufficient and must be expanded. Specifically, we introduce a topology-based decision mechanism that analyzes the connectivity of the pose-graph to determine whether the system should continue operating in localization mode or activate mapping to incorporate new observations.}
Furthermore, our approach does not specifically require the prior models to be generated using our SLAM algorithm, as is the case in~\cite{zou2024lta}.

\subsection{Pose-graph Connectivity as Quality Measure}

The use of topological information has recently gained interest in the robotics community, especially since mo\-dern detailed and rich representations call for hierarchical and efficient ways of measuring the quality of the SLAM estimates.
The impact of the connectivity of PGs into the estimation problem is well-known~\cite{bailey2006simultaneous}. In~\cite{khosoussi2019reliable}, the intuition behind this link
is first formalized relating different graph connectivity indices to estimation reliability.
Placed and Castellanos~\cite{placed2022general} relate the so-called \emph{optimality criteria} to the weighted connectivity of PGs in which edges are weighted with a scalar function of the Fisher Information Matrix (FIM) they encode (\eg~its trace).
This allows to efficiently make an uncertainty-aware assessment of the quality of the SLAM estimates, with direct application to active perception.
Despite growing interest, these metrics have not yet been used to inform multi-session mapping.

%% file: method_iros.tex
\section{Method}

\begin{figure}[!t]
    \centering
    \includegraphics[
        width=\linewidth,
        keepaspectratio
    ]{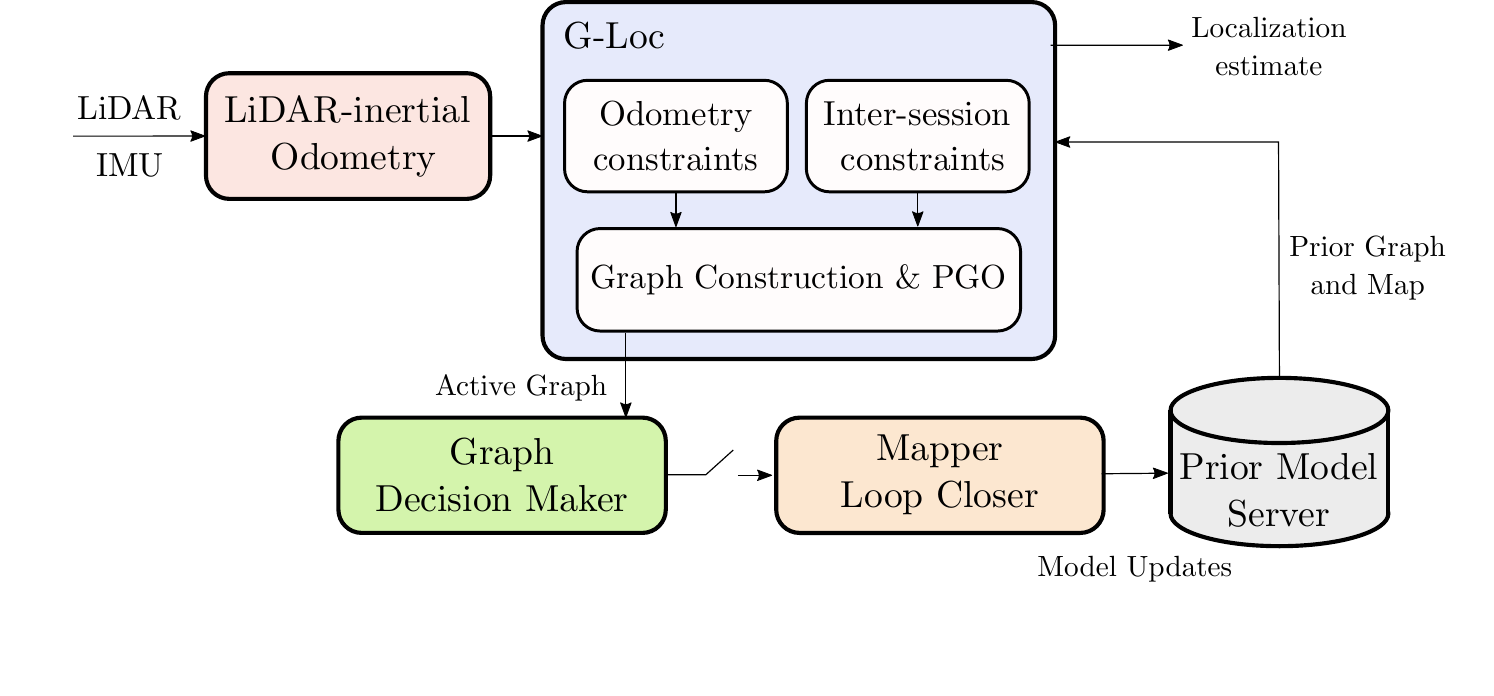}
    \caption{Overview of the proposed method.}
    \label{fig:overview}
\end{figure}

\changes{We propose a topology-driven framework for multi-session robotic operation that enables automatic switching between map-based localization and mapping.}
It features a novel decision-making module that evaluates the connectivity of the underlying joint PG
\changes{formed by the current session and the prior map to determine whether the available structural information is sufficient to support reliable localization.}
\changes{The system integrates 3 main components: (i) a map-based localization backbone, (ii) a mapping and}
a robust LC module that
searches for both intra- and inter-session constraints,
\changes{and (iii) a topology-driven decision mechanism.}
See~\cref{fig:overview}.

\begin{figure*}
    \centering
    \includegraphics[width=0.7\linewidth]{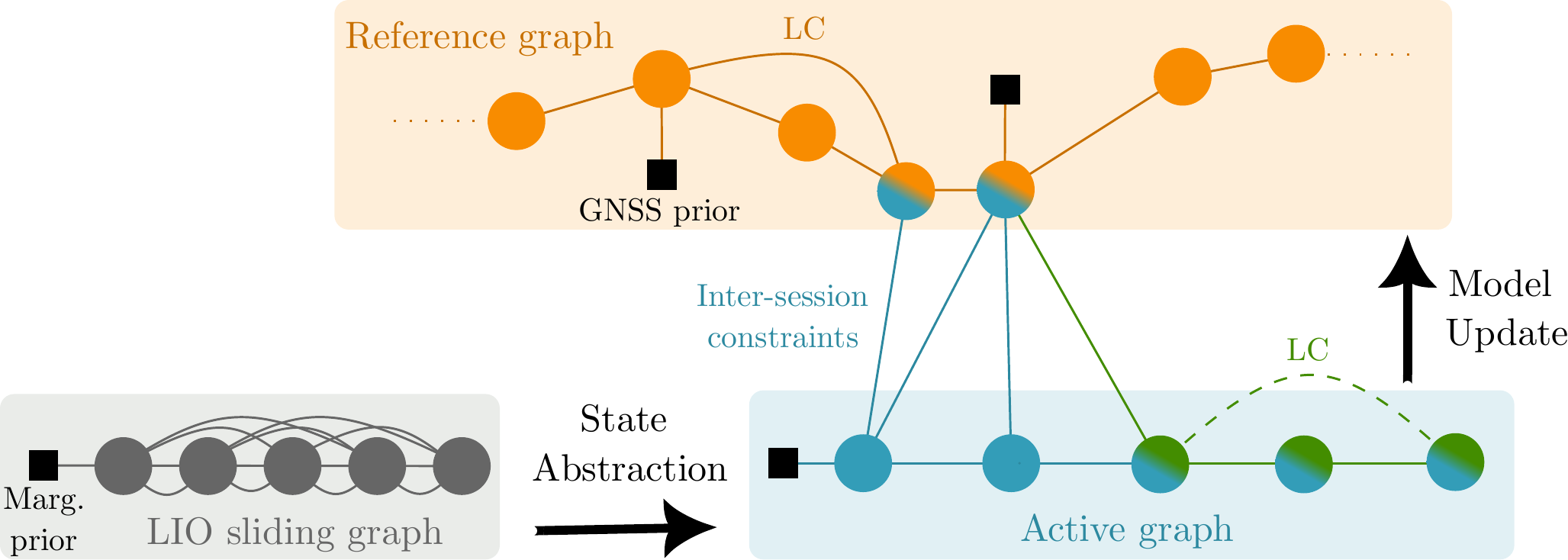}
    \caption{Visualization of the graphs used in the multi-session \changes{framework}. The active graph (blue) is \changes{built} from selected key frames \changes{of} the odometry graph (gray) \changes{and} their condensed measurements, \changes{together with} vertices from the reference graph (orange) connected via inter-session correspondences. When triggered by the topological decision-making module (\textit{e.g.},~in unmapped regions), a subset of the active graph (candidate graph, green) is integrated into the reference model. LC and global PGO are performed to 
    ensure consistency.}
    \label{fig:graphs}
\end{figure*}

\subsection{Multi-session Graph Localization}


\changes{We adopt}
G-Loc~\cite{Montano-OlivanRAL2024} \changes{as the localization backbone, which} 
combines LiDAR-inertial Odometry (LIO) and map matching constraints into a multiple graph optimization framework. 
~\cref{fig:graphs} depicts the different graphs involved in the localization process. 
First, LIO hyper-edges and \changes{scan-to-scan} matching are optimized within a sliding-window fashion (gray graph in that figure), producing coarse estimates of the robot poses at high frequency (up to IMU rate). A marginalization prior is used to retain information from past states outside the optimization window.

The \emph{active graph} (blue) contains vertices that encode a sparse subset of the LIO states, \textit{i.e.,}~LIO-optimized poses of the most recent $n$ key frames, {\small$\mathcal{T} \triangleq \{\mathbf{T}_1, \dots, \mathbf{T}_n\mid\mathbf{T}_i\in SE(3)\}$}, together with their associated submaps. Such vertices are connected consecutively via submap-to-submap matching constraints.
In addition, this graph includes inter-session constraints (submap-to-map matching) that link {\small$\mathcal{T}$} with nearby vertices of the \emph{reference graph} (orange) and their $k$-neighbors.
The active graph is optimized to obtain a precise localization, keeping the nodes from the reference graph fixed, \textit{i.e.,}
\begin{gather}
    \mathcal{T}^\star = 
    \underset{\mathcal{T}}{\mathrm{argmin}} \Bigg(     
        \textstyle \sum\limits_{(i,j)\in\mathcal{E}_\text{intra}} \| \boldsymbol{e}_{i,j}^{\text{intra}}\|^2
        + \sum\limits_{(i,r)\in\mathcal{E}_\text{inter}} \| \boldsymbol{e}_{i,r}^{ \text{inter}}\|^2 \Bigg)
        \,,
         \label{eq:optimization}
\end{gather}
\noindent where {\small$\|\boldsymbol{e}\|^2\triangleq (\boldsymbol{e}^T  \boldsymbol{\Sigma}^{\texttt{-}1}  \boldsymbol{e})\in\mathbb{R}$} is the quadratic error for a measurement with covariance {\small$\boldsymbol{\Sigma}$}, {\small$\boldsymbol{e}_{i\texttt{-}1, i}^{\text{intra}}\in\mathfrak{se}(3)$} are the residuals associated with consecutive submap matching (intra-session constraints), and
{\small$\boldsymbol{e}_{i,r}^{\text{inter}}\in\mathfrak{se}(3)$} are those associated with the correspondences between current observations and the reference map (inter-session constraints). The pair set {\small$\mathcal{E}_{\text{intra}}\triangleq\{(i,j)\mid1\leq i<j\leq n, j=i+1\}$} and {\small$\mathcal{E}_{\text{inter}}\triangleq\{(i,r)\mid 1\leq i\leq n, r\in\mathcal{I}_\text{ref}\}$}, with $\mathcal{I}_{\text{ref}}$ the index set of vertices that belong to the reference graph.

Solving~\eqref{eq:optimization} enables limited spatial and topological association with previous sessions.
G-Loc treats the reference model (PG and point cloud submaps) as a passive input ---immutable information used solely for localization. 
This approach underutilizes the rich information encoded in the PG, including its representation of uncertainty.
To address this, we extend~\cite{Montano-OlivanRAL2024} by transforming the reference model into an active component of the localization pipeline: when the system traverses an unseen or poorly mapped area, the model is incrementally updated with new observations. See the bottom stream in~\cref{fig:overview}.

First, all vertices together with their associated submaps that leave the sliding window of key frames ({\small$\mathcal{T}$}) are retained.
To bound memory usage, temporal and graph-distance limits are imposed, pruning elements that exceed them.
When the mapping mode is triggered by the decision-making module (described in~\cref{S:decision_making}), these inactive nodes are selected to form the \emph{candidate graph} (green in~\cref{fig:graphs}), which is then merged into the reference model.
As they enter the reference graph, they are automatically incorporated into a LC database. Robust LC search, matching \changes{and false-positive rejection (see~\cite{Montano-OlivanROBOT24})} occurs in a separate thread, improving the overall connectivity and accuracy.
Therefore, under this mapping mode, the optimization problem~\eqref{eq:optimization} becomes:
\begin{align}
    \{\mathcal{T}, \mathcal{R}\}^\star = 
    \underset{\{\mathcal{T},\mathcal{R}\}}{\mathrm{argmin}}   \Bigg( & \textstyle \sum\limits_{(i,j)\in\mathcal{E}_{\text{intra}}} \| \boldsymbol{e}_{i,j}^{\text{intra}}\|^2
    + \sum\limits_{(i,r)\in\mathcal{E}_{\text{inter}}} \| \boldsymbol{e}_{i,r}^{ \text{inter}}\|^2 \nonumber \\
    &\textstyle + \sum\limits_{(r,s)\in\mathcal{E}_{\text{LC}}} \| \boldsymbol{e}_{r,s}^{ \text{LC}}\|^2 \Bigg) \,, 
    \label{eq:optimization2}
\end{align}
\noindent where {\small$\mathcal{R}\triangleq\{\mathbf{T}_r\in SE(3)\mid r\in\mathcal{I}_{\text{ref}}\}$} denotes the set of reference poses, which in mapping mode can grow over time and are jointly optimized together with {\small$\mathcal{T}$}. Note that global PGO is now performed, taking into account both the current and reference poses.
In addition, {\small$\boldsymbol{e}_{r,s}^{\text{LC}}\in\mathfrak{se}(3)$} are the residuals from LC constraints and {\small$\mathcal{E}_\text{LC}\triangleq \{(r,s)\mid r,s\in\mathcal{I}_\text{ref}, r\neq s\}$}.

\subsection{Decision Making}
\label{S:decision_making}


The decision of whether to localize on the previous map or extend it is guided by the topology of the active and reference graphs. 
Intuitively, if these two graphs are disconnected, this suggests that either the system is entering an unvisited area or the overlap between the observations is insufficient ---both scenarios where mapping becomes necessary. Moreover, the system can identify regions with highly uncertain connections, in which case reinforcing the map is also desirable.

\subsubsection{On the Graph Signatures}
The structural patterns, or \emph{signatures}, of PGs reflect the conditioning of the underlying optimization problem and can thus serve as  measures of the estimation accuracy~\cite{khosoussi2019reliable}.
For instance, well-distributed and broadly spanning loops enhance global consistency, whereas sparse and weakly connected structures can be related to drift, poor conditioning and solution instability.
The Laplacian matrix of a graph, {\small$\mathbf{L}\in \mathbb{R}^{m\times m}$} with {\small$m$} the amount of vertices, provides a compact algebraic representation of its topology. In particular, the spectrum of {\small$\mathbf{L}$} encodes fundamental structural properties, with eigenvalues {\small$\lambda = (\lambda_1=0, \lambda_2, \ldots, \lambda_m)$} when the graph is connected.
Among the various metrics derived from spectral graph theory, the average node degree, {\small$d$}, is one of the simplest and most widely used. It reflects the mean connectivity and can be formally defined as:
\begin{equation}
    d \triangleq \frac{1}{m}\textstyle\sum\limits_{k=2}^m \lambda_k \, .
\end{equation}
Intuitively, {\small$d$} just captures the ratio between the number of observations and the number of optimization variables (\textit{i.e.,}~robot poses). Consequently, PGs with higher ratios yield larger values of {\small$d$}, which correspond to denser connectivity and, in general, more reliable estimation performance.
Another fundamental metric is the Fiedler value, {\small$\lambda_2$}, \textit{i,e,.}~the second-smallest eigenvalue of {\small$\mathbf{L}$}. It reflects how well-connected a graph is: it is strictly positive if and only if the graph is connected, and larger values imply stronger connectivity and improved robustness against partitioning.

Nevertheless, such metrics offer only a limited view of estimation accuracy. For instance, two graphs with same number of edges and vertices yield the same value of {\small$d$}, regardless of whether the edges connect consecutive nodes or form loops. Moreover, the previous indices do not capture uncertainty in the observations.
To overcome this limitation, the use of \emph{weighted} connectivity indices has been proposed~\cite{placed2022general}, relating them to optimality criteria. By assigning each edge a weight based on a measure of the uncertainty it encodes (\textit{e.g.,}~trace, minimum eigenvalue), one can compute the weighted node degree and the weighted Fiedler value, which approximate T- and E-optimality, respectively.

\subsubsection{Two-step Consensus}
Topological metrics of the PG \changes{provide a compact representation of the structural reliability of the PG and therefore serve as} useful indicators of the graph connectivity and the estimates accuracy. By tracking their evolution over time, it is possible to anticipate disconnections and determine weakly connected and uncertain regions, \textit{i.e.,}~when
the reference model needs to be updated.
Let us consider the joint PG formed by the active and reference graphs (blue and orange in~\cref{fig:graphs}) from which we can generate two weighted graphs where edges are weighted with the trace and minimum eigenvalue, respectively, of the Information matrix encoded by each edge. The first graph will be used to compute the weighted connectivity index {\small$\bar{d}$}, and the second to compute {\small$\bar{\lambda}_2$}. 
The objective of the two-step consensus strategy is to enable automatic switching between mapping and map-based localization modes based on the evolution of {\small$\bar{d}$} and {\small$\bar{\lambda}_2$}.
\changes{The two-step consensus mechanism prevents spurious mapping triggers caused by transient localization errors, ensuring that mapping is activated only when the prior model becomes structurally insufficient.}

\begin{figure*} [t!]    
    \centering
    \includegraphics[ width=0.60\linewidth]{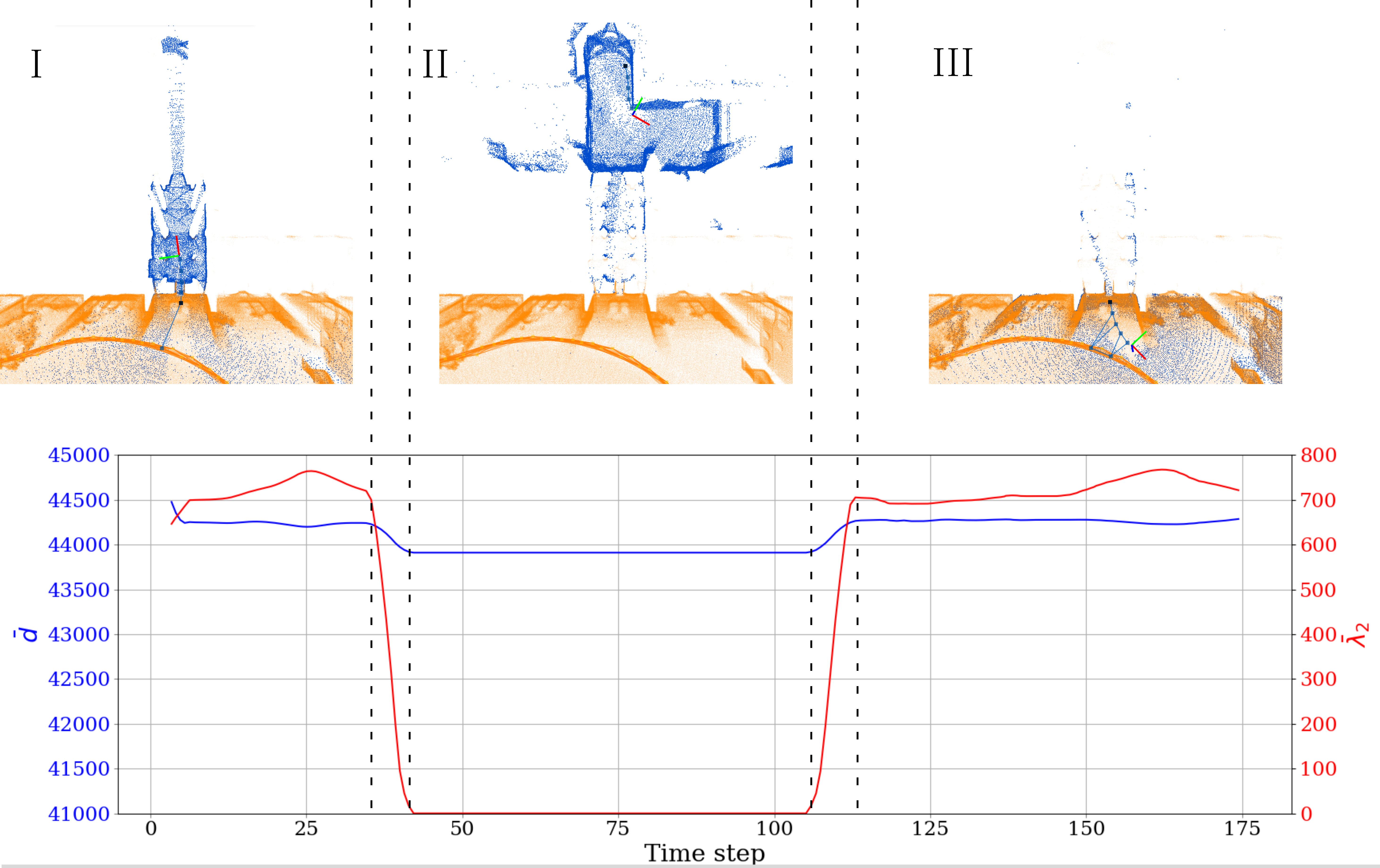}
    \caption{Illustration of the evolution over time of the weighted connectivity indices (blue: $\bar{d}$, red: $\bar{\lambda}_2$) during the traversal of previously unmapped regions. \emph{Stage I} corresponds to traversal within previously mapped regions, where localization is performed. \emph{Stage II} indicates entry into unmapped regions, resulting in disconnection between the reference and active graphs, and necessitating mapping. \emph{Stage III} marks reconnection with the prior model and re-establishes map-based localization. The topmost images illustrate examples of these phases: the orange map and PG are the reference model, while the blue ones correspond to the online point cloud and sliding graph.}
    \label{fig:t_e_opt}
\end{figure*}

Given the quick computation of {\small$\bar{d}$}, we first evaluate its deviation w.r.t. its moving average. Using the moving average rather than a fixed value allows to account for the specifics of the prior model, the algorithm configuration, the current observations, etc. If {\small$\bar{d}$} deviates more than {\small$2\sigma$} from the mean, we compute {\small$\bar{\lambda}_2$}. 
If {\small$\bar{\lambda}_2=0$}, we immediately trigger mapping, since this condition indicates a disconnection between the active and reference graphs (due to matching failures caused by either poorly mapped regions or the presence of new areas).
Furthermore, if {\small$\bar{\lambda}_2> 0$} but {\small$\bar{d}$} decreases beyond {\small$4\sigma$}, we also trigger mapping. This situation is related to weakly connected regions or edges carrying limited information, cases in which rein\-forcing the prior model is also desired.
Conversely, while in mapping mode, an increase of {\small$\bar{d}$} back within the confidence bounds together with a recovery of {\small$\bar{\lambda}_2>0$} indicates the re-establishment of reliable structural information, thereby triggering a switch to map-based localization. To prevent weak inter-session constraints, we enforce this two-step validation rather than relying solely on {\small$\bar{\lambda}_2>0$}.
As the triggering mechanism is based on the multiplicity of the zero eigenvalue of the graph Laplacian, disconnection is detected exactly rather than probabilistically, guaranteeing precision and recall by construction.

We present \cref{fig:t_e_opt} to illustrate the evolution of the topological metrics that drive the decision-making mechanism. We use two partially-overlapping sequences of the New College datasets \cite{ramezani20, zhang21}. For the purpose of analyzing the behavior of the weighted indices, mapping mode is not triggered ---focus solely on when it should be activated and deactivated. A reference map (teal point cloud in that figure) of an outdoor region is provided to the algorithm. In stage I, the robot navigates within the mapped region, maintaining stable connectivity indices determined by the sliding graph (blue) configuration and the reference graph (orange) connectivity. 
At some point, the robot exits the outdoor region and enters an unknown indoor corridor (upper part in~\cref{fig:t_e_opt}). 
As \changes{it} transitions from stage I to II, the sliding graph gradually disconnects from the prior model: both {\small$\bar{d}$} and {\small$\bar{\lambda}_2$} decrease. 
During II, the graph remains disconnected (as mapping is not activated), with {\small$\bar{\lambda}_2=0$}.
Finally, the robot returns to the mapped region, transitioning to stage III and reconnecting the sliding graph with the prior model: {\small$\bar{d}$} increases to values akin to those before disconnection and {\small$\bar{\lambda}_2>0$}.



\begin{figure*}
    \centering
    \includegraphics[width=0.58\linewidth]{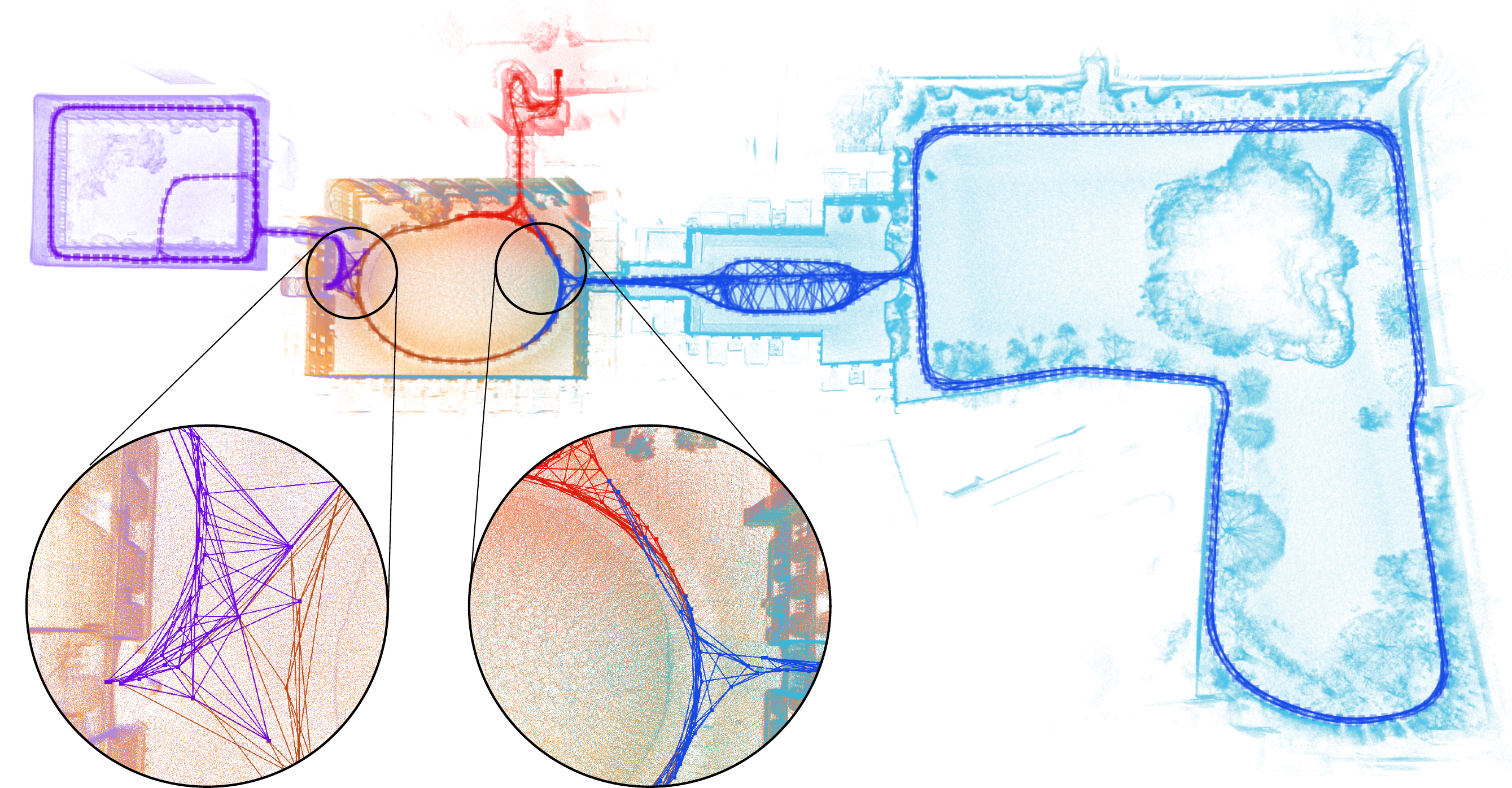}
    \caption{Multi-session mapping results in NC and NCE datasets: \textit{Quad-e} (orange), \textit{Quad-m} (teal), \textit{Quad-h} (red), \textit{Cloister} (purple), and \textit{Short} (blue). Notably, overlapping regions are modeled only once (both in the map and the PG).}
    \label{fig:nc_mapas}
\end{figure*}

\begingroup
\setlength{\tabcolsep}{8pt}
\renewcommand{\arraystretch}{1.3}
\begin{table*}[t!]
    \centering
    \small
    \begin{tabular}{l|r||c|c|c|c|c||c}
        \multicolumn{2}{c||}{\textbf{Method / Sequence}} & \textit{Quad-e} & \textit{Quad-m} & \textit{Quad-h} & \textit{Cloister} & \textit{Short} & Avg. \\ \hline
        \multirow{3}{*}{SLAM} 
        & LIO-SAM~\cite{shan2020lio} &\underline{0.074} & \textbf{0.067} & 0.13 & \textbf{0.074} & \underline{0.41} & \underline{0.15}\\
        & NV-LIOM~\cite{chung24} &0.076 & 0.076 & 0.18 & 0.077 & 0.48 & 0.18 \\
        & LG-SLAM~\cite{Montano-OlivanROBOT24} & \textbf{0.066} & \underline{0.072} & 0.090 & \underline{0.076} & \textbf{0.38} & \textbf{0.14 }\\ 
        \hline
        \multirow{2}{*}{Map-based Localization} & HDL-Loc~\cite{koide17} & -- & 0.10 & $\times$ & $\times$ & $\times$ & $\times$ \\ 
        & G-LOC~\cite{Montano-OlivanRAL2024} & -- & \textbf{0.067} & \underline{0.087} & 0.21 &1.61 & 0.39 \\ \hline 
        \multicolumn{2}{c||}{\emph{Ours}} & \textbf{0.066} & \textbf{0.067} & \textbf{0.082} & 0.077 & 0.43 & \textbf{0.14} \\
    \end{tabular}
    \caption{ATE RMSE (m) in NC and NCE dataset. SLAM methods build separate maps for each sequence. Map-based localization methods employ \textit{Quad-e} as reference map. Our method incrementally concatenates sessions (left to right). Best results are bold and second best are underlined. $\times$ indicates an unrecoverable failure.}
    \label{tab:nce_benchmark}
\end{table*}
\endgroup

\changes{
\subsection{Initial Inter-Session Alignment}
To initialize localization in a prior map, we consider two complementary strategies. 
First, we include a place-recognition approach based on ScanContext++~\cite{kim2022scancontext} retrieves the $k$-best loop candidates from the prior map, followed by a 3D alignment step to select the most consistent one and obtain an initial pose estimate within the existing map. Second, global localization (\eg~GNSS measurements) can be seamlessly integrated into the SLAM framework to obtain a globally consistent prior that supports direct localization. 

\changes{}
}

%% file: experiments.tex
\section{Experiments}

In order to validate the proposed method, we have conducted two sets of experiments. 
The first utilizes two widely used datasets to evaluate the performance of our system in a controlled setting, enabling comparison against existing approaches and a detailed analysis.
The second experiment involves a real-world deployment in an underground mine, aimed at demonstrating the system's robustness and application in challenging environments. 
All experiments were performed on a system equipped with an Intel Core Ultra 7 155H processor and an NVIDIA GeForce RTX 4070 GPU.

A shared requirement for all experiments is the availability of prior topological and geometric models of the environment (\textit{i.e.,}~a PG and a point cloud).
To fulfill this requirement, we use LG-SLAM~\cite{Montano-OlivanROBOT24}, a robust and accurate framework for LiDAR-inertial mapping.

\subsection{Dataset Evaluation}

The Newer College (NC)~\cite{ramezani20} and Newer College Extension (NCE)~\cite{zhang21} datasets contain 
data recorded using a handheld device in diverse environments.
In \changes{NC,} 
data was recorded using an Ouster OS1-64 LiDAR and its internal Inertial Measurement Unit (IMU)
\changes{while NCE uses}
an Ouster OS0-128 and an Alpasense Core 
\changes{Both}
datasets \changes{provide}
ground-truth trajectories
\changes{and include}
sequences with partially overlapping trajectories 
that have been previously used for multi-session SLAM evaluation~\cite{hu2024ms,menezes2023multisession}. 

The experiment consisted of gradually processing five overlapping sequences from NC and NCE.
First, we create a base model with \emph{Quad-e} (orange in~\cref{fig:nc_3maps,fig:nc_mapas}). Then, we perform incremental updates using the \changes{\emph{Quad-h}} (teal), \emph{Cloister} (purple), and \emph{Short} (blue) sequences, each of which extends the information provided by the preceding one.~\cref{fig:nc_mapas} illustrates the joint model after processing all sequences. Each new region is shown in a different color, showcasing the contribution of each session, the strong inter-session links, and the fact that overlapping regions are modeled only once. 

To evaluate the performance of the proposed method, we present the Absolute Trajectory Error (ATE) Root Mean Squared Error (RMSE) for each individual sequence, and compare our approach with state-of-the-art LIO/SLAM~\cite{Montano-OlivanROBOT24,shan2020lio, chung24} methods (which build separate maps for each sequence) and map-based localization systems~\cite{Montano-OlivanRAL2024,koide17} (all of them using \emph{Quad-e} as reference map). 
Note that, in contrast, our method concatenates the sessions and incrementally leverages the information from previous sequences.
The results are reported in~\cref{tab:nce_benchmark}.
Overall, the proposed approach produced the best results for three out of five sequences and had the lowest average RMSE, alongside LG-SLAM.
Since the mapping module is similar to that of LG-SLAM, both approaches achieved the same level of accuracy in \textit{Quad-e}, where no prior information was available. 
Similarly, when the sequences fully overlap (\ie~in \textit{Quad-m}), our method exhibits the same behavior as G-Loc.
However, in \textit{Quad-h}, where there is a new indoor region not present in the previous sequences, our method outperforms all others. As this new area is small, G-Loc still performs accurately, demonstrating its robustness even in partially unmapped regions.
However, map-based localization methods fail or accumulate drift in \textit{Cloister} and \textit{Short}, as there are large unvisited regions. Notably, our approach achieves accuracy comparable to SLAM methods while substantially reducing graph complexity by alternating between localization and mapping modes, and providing a joint model.

\begin{figure}
    \centering
    \includegraphics[width=0.7\linewidth]{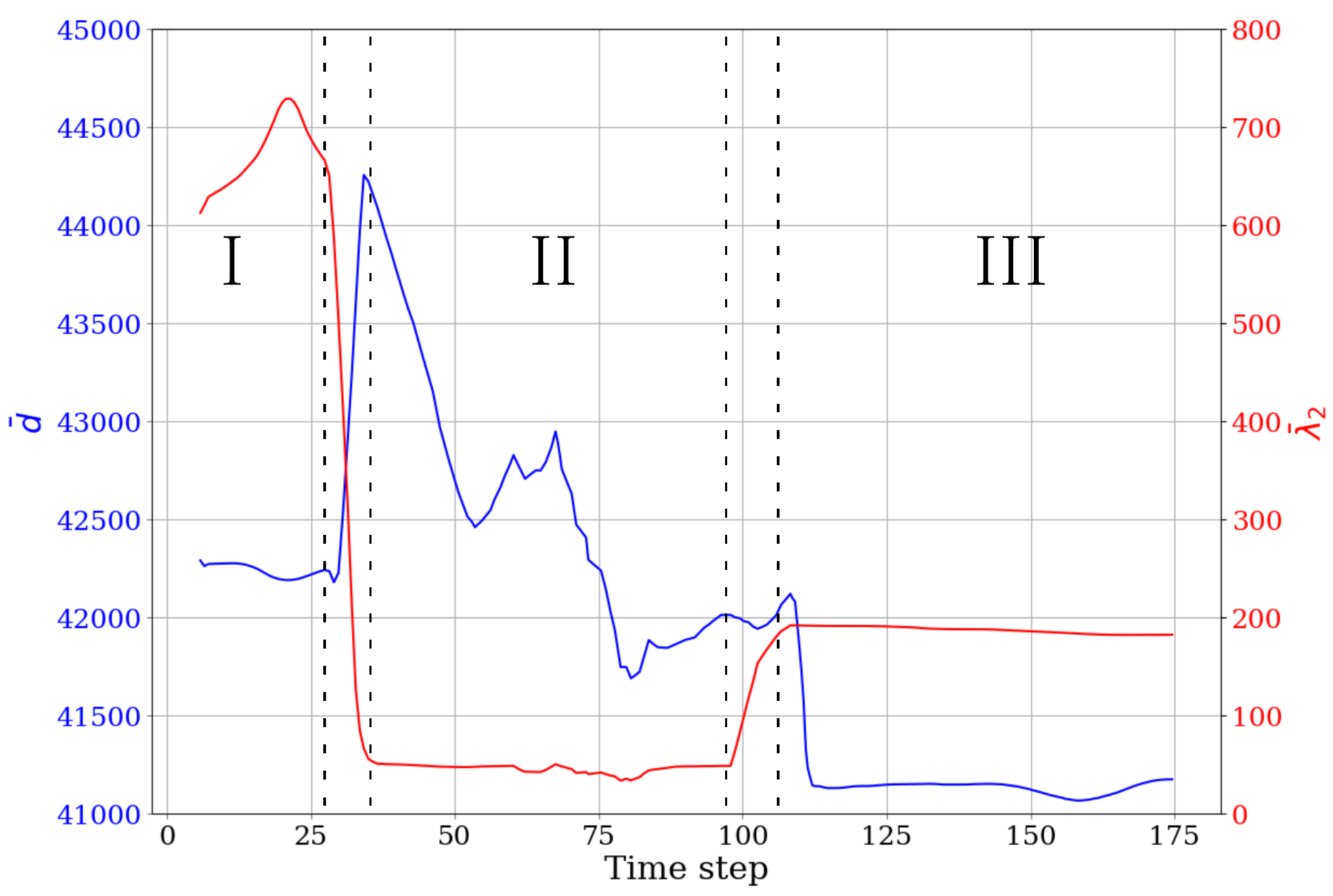}
    \caption{Evolution of the weighted average node degree (blue) and Fiedler value (red)   over time while processing \textit{Quad-h} sequence.}
    \label{fig:opt_evol}
\end{figure}

\begin{figure}
    \centering
    \includegraphics[width=0.62\linewidth]{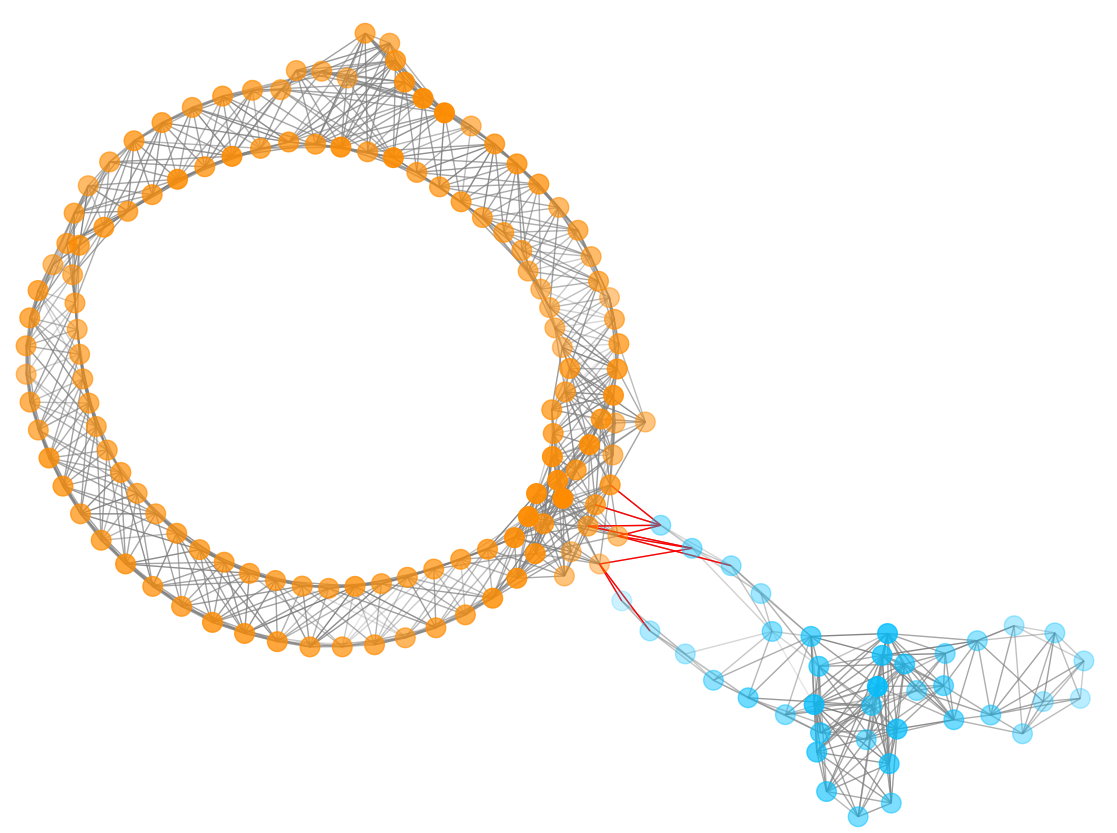}
    \caption{Visualization of the joint PG after processing \emph{Quad-e} and \emph{Quad-h}. Orange nodes belong to the prior graph (\emph{Quad-e}), while blue ones are the new nodes incorporated to the model during \emph{Quad-h} sequence. Red edges represent the weakest part of the graph.} 
\label{fig:graph_networkx}
\end{figure}

\cref{fig:opt_evol} presents the evolution of graph weighted connectivity indices while processing \emph{Quad-h} sequence (using \emph{Quad-e} as prior model). This figure is similar to~\cref{fig:t_e_opt}, but now we analyze the behavior during operation (\ie~mapping mode is triggered).
At the transition from I to II, the mapping/LC modules establish multiple connections between the active and reference graphs, leading to a slight increase in {\small$\bar{d}$}. At the same time, these connections ensure that the two graphs remain connected (thus {\small$\bar{\lambda}_2>0$}). The sharp drop in the \changes{Fiedler} value is explained by the fact that the newly explored region is largely separated from the prior model, being physically connected only through a narrow corridor, which reduces the graph's robustness against partitioning.
This effect is also illustrated in~\cref{fig:graph_networkx}, where the red edges highlight the weakest structural links between the two sessions. The prior graph is shown in orange, and the newly explored regions in blue; also, node transparency reflects per-node value of {\small$\bar{d}$}.
The transition from II to III corresponds to the reconnection phase. On the one hand, the weighted average node degree stabilizes to a value comparable to stage I, with a slight decrease ({\small$\approx2\%$}) attributed to the covariance of the measurements in the new region.
On the other hand, {\small$\bar{\lambda}_2$} gradually increases as the links between the two sessions are reinforced, until reaching a plateau. 
Compared to stage I, the lower plateau value indicates the presence of two more disjoint regions in the graph. 
In this experiment, the new area is connected to the prior model only through a narrow corridor, which reduces the overall connectivity relative to the strongly connected 
reference model (see~\cref{fig:graph_networkx}).

\cref{tab:ms_results} contains a comparison against a multi-session systems in \emph{Parkland} (S1) and \emph{Short} (S2) sequences from~\cite{ramezani20}. We compare against MS-Mapping~\cite{hu2024ms} and two multi-session localization algorithms (M2F and F2F) reported in~\cite{hu2024ms}. Specifically, M2F localizes the new session point cloud frame-by-frame on the prior map to establish constraints, while F2F localizes frames from the current session against the joint map.
The results of M2F and F2F are obtained directly from~\cite{hu2024ms}.
The first column reports ATE RMSE after processing each sequence individually, where our approach outperforms the alternatives. The second column shows results when leveraging cross-session information ---using \emph{Parkland} while processing \emph{Short}, and vice versa.
Results for the combined sequences show comparable results, with lower computational requirements.

\begingroup
\begin{table}[t]
    \centering
    \begin{tabular}{l||c c|c c}
        \multirow{2}{*}{\textbf{Method / Seq.}} & \multicolumn{2}{c|}{Individual} & \multicolumn{2}{c}{Combination} \\
         & \textit{S1} & \textit{S2} & \textit{S1} (given \textit{S2}) & \textit{S2} (given \textit{S1}) \\
        \hline
        M2F (from~\cite{hu2024ms}) & 0.28 & {0.46} & \underline{0.27} & 0.49  \\
        F2F (from~\cite{hu2024ms}) & 0.28 & {0.46} & \underline{0.27} & 0.50  \\
        MS-Mapping~\cite{hu2024ms}  & \underline{0.26} & \underline{0.45} & 0.29 & \textbf{0.40}  \\ \hline
        \textit{Ours}  & \textbf{0.17} & \textbf{0.38} & \textbf{0.24} & \underline{0.47}  \\
    \end{tabular}
    \caption{Comparison of ATE RMSE (m) for multi-session approaches on \textit{Parkland} (S1) and \textit{Short} (S2) sequences.}
    \label{tab:ms_results}
\end{table}
\endgroup

Finally, regarding the execution time, our approach substantially reduces graph complexity by alternating between localization and mapping modes (\textit{e.g.,} the number of nodes of the combined S1 and S2 is 600, whereas the individual S1+S2 is 770). This reduction directly translates into improved computational efficiency, highlighting the benefits of leveraging multi-session information without remapping entire sequences. On average, the proposed method required $50$ ms to process each keyframe and $196$ ms for LC and PGO. In comparison, the SLAM baseline~\cite{Montano-OlivanROBOT24} required $48$ ms to process each keyframe and $231$ ms for LC and PGO. Notably, in our approach, the LC thread was active only during a part of the sequences, further reducing the overhead.



\subsection{Real-World Experiment}

The Kemi Mine is the largest underground mine in Findland, with $1$ km depth. The experiments covered a distance of $8$ km along a series of spiral primary and secondary galleries.
Using a prototype vehicle and following diverse trajectories throughout the tunnels, data from the mine was collected using two Robosense M1 solid state LiDARs \changes{(front- and rear-facing)} and an industrial-grade IMU. \changes{Both LiDARs were used to build the prior maps while only the front-facing LiDAR was used for localization to demonstrate the performance with mismatched field-of-view. Although the standard configuration uses both sensors, this experiment demonstrates that reliable localization can still be achieved under these conditions.}
In addition, Wi-Fi localization was used to provide a rough location estimate and correctly initialize the map-based localization algorithm, thanks to the large amount of access points located throughout the tunnels.

\begin{figure}
    \centering
    \includegraphics[width=0.65\linewidth]{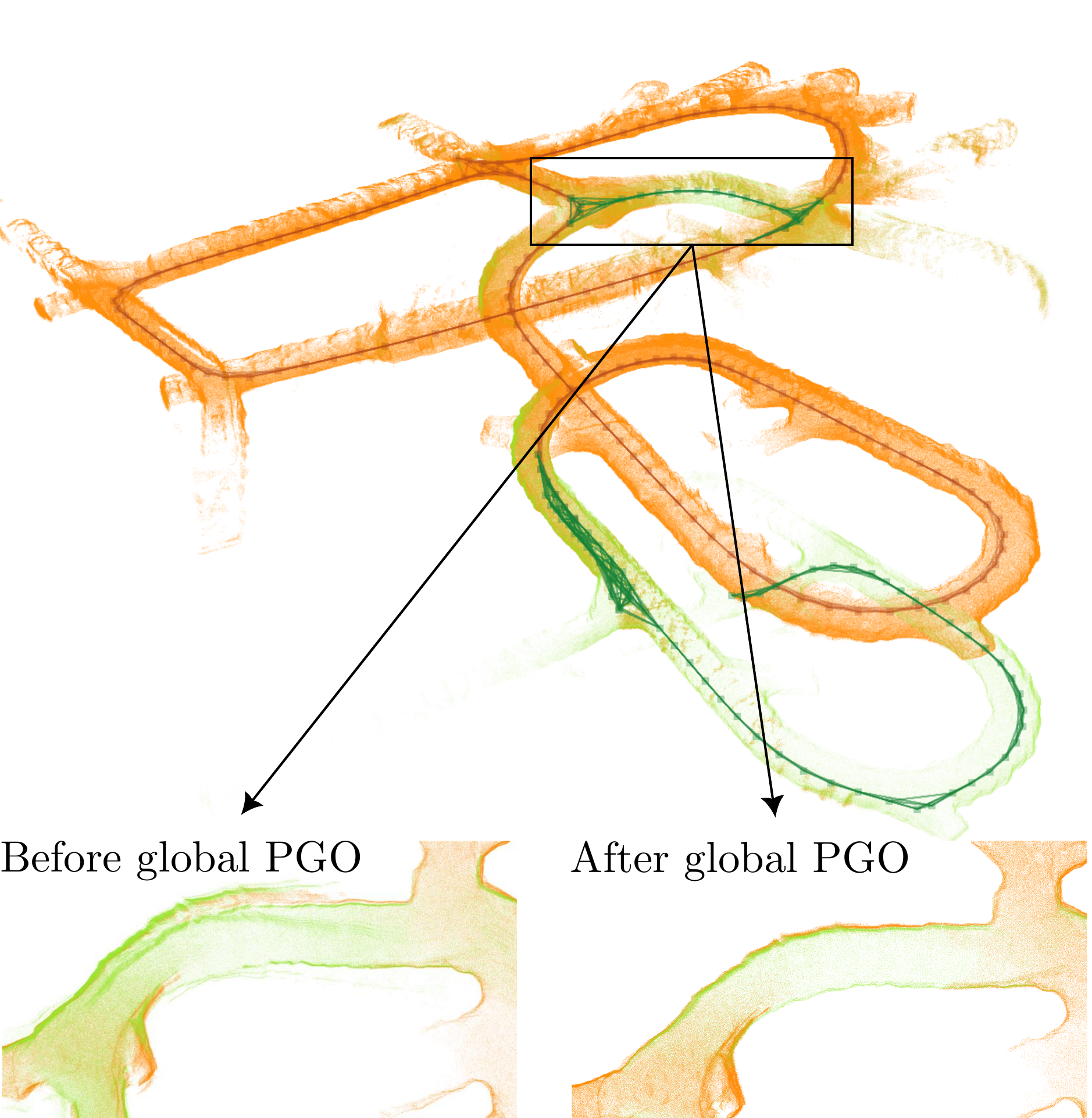}
    \caption{Multi-session results in the Kemi underground mine. The boxed region shows a loop closure enforced between two areas previously mapped but disconnected. Furthermore, the zoomed-in images further illustrate the impact of global PGO.}
    \label{fig:kemi}
\end{figure}

\cref{fig:kemi} contains the resulting map for two of the sequences in the mine, which partially overlap. The prior SLAM-generated model is shown in orange, while the extended model from the second session appears in green.
Two areas of the prior map are seamlessly expanded into previously unmapped regions. The PG constructed during the second session exhibits strong internal connectivity, as well as robust links to the prior graph ---enabled by the intelligent module responsible for orchestrating mapping and LC operations.
Notably, one of the new regions (boxed area in~\cref{fig:kemi}) enforces a loop in the prior map. This clearly improves the overall connectivity of the joint PG. Also, thanks to the global optimization performed where uncertainties of both mapping processes are accommodated, the consistency and precision of the point cloud representation are enhanced.

Specifically, the average node degree of the prior graph (orange graph in~\cref{fig:kemi}) is $5.6$. This value increased to $6.5$ in the updated graph after processing the second sequence with our multi-session approach (orange and green graphs in the 
figure).
The analysis of the {\small$\bar{d}$} is equivalent ($2184.6$ for the first case, and $2719.9$ for the second), indicating the presence of strong inter-session links with low-uncertainty.
The weighted number of spanning trees, a measure of global graph connectivity~\cite{khosoussi2019reliable, placed2022general}, remained nearly constant in both cases ($1.4$ and $1.3$, respectively); this shows that the links between the two sessions did not introduce any significant \changes{connectivity} degradation.
The analysis of the eigenvectors for both graphs reveals that the weakest region lies within the first session in both cases; \textit{i.e.,}~the new session did not introduce any weaker edges.
Regarding time consumption, executing SLAM required, on average, $59.9$ 
ms to process each frame (including odometry and mapping). In addition, $149.6$ ms were consumed by the LC thread per iteration (including PGO). These values remained similar for both sequences. In contrast, 
our multi-session approach required $45.3$ 
ms to process each frame. Mapping, LC and global PGO were selectively triggered, requiring $109.9$ ms on average but being active just $10\%$ of the time.

%% file: conclusions.tex
\section{Conclusions}

\changes{In this work, we proposed a topology-driven framework for multi-session robotic operation that determines when map-based localization should transition to mapping.
By leveraging connectivity metrics derived from spectral graph theory, the proposed method provides a principled mechanism for detecting when map expansion is required while avoiding redundant remapping of previously visited areas.
Experiments on public datasets and real-world deployments demonstrate that the approach achieves localization accuracy comparable to state-of-the-art SLAM systems and enables multi-session operation.}
Future work will focus on efficient, real-time change detection on previously visited areas and its integration into the existing model, strengthening the PG structure when required and enhancing overall mapping.